\pdfoutput=1
\documentclass[11pt]{article}

\usepackage[preprint]{latex/acl}
\usepackage{amsmath, amssymb}
\usepackage{booktabs}
\usepackage{times}
\usepackage{latexsym}
\usepackage{subcaption,seqsplit}
\usepackage{wrapfig}

\usepackage[T1]{fontenc}
\newcommand{\wrapt}[1]{\texttt{\seqsplit{#1}}}

\usepackage[utf8]{inputenc}

\usepackage{tipa}
\usepackage{microtype}

\usepackage{inconsolata}

\usepackage{graphicx}
\usepackage{todonotes}

\title{Energy Considerations of Large Language Model Inference and Efficiency Optimizations}

\author{
 \textbf{Jared Fernandez\thanks{Equal contribution}\textsuperscript{1}},
 \textbf{Clara Na\footnotemark[1]\textsuperscript{1}},
 \textbf{Vashisth Tiwari\footnotemark[1]\textsuperscript{1}}, \\
 \textbf{Yonatan Bisk\textsuperscript{1}},
 \textbf{Sasha Luccioni\textsuperscript{2}},
 \textbf{Emma Strubell\textsuperscript{1}}
\\
\\
 \textsuperscript{1}Carnegie Mellon University,
 \textsuperscript{2}Hugging Face,
\\
 \small{
   \textbf{Correspondence:} \{\href{mailto:jaredfern@cmu.edu}{jaredfern}, \href{mailto:clarana@cmu.edu}{clarana}, \href{mailto:vashisthtiwari@cmu.edu}{vashisthtiwari}\}@cmu.edu}
 }

\begin{document}
\maketitle

\begin{abstract}

    As large language models (LLMs) scale in size and adoption, their computational and environmental costs continue to rise.  Prior benchmarking efforts have primarily focused on latency reduction in idealized settings, often overlooking the diverse real-world inference workloads that shape energy use. In this work, we systematically analyze the energy implications of common inference efficiency optimizations across diverse Natural Language Processing (NLP) and generative Artificial Intelligence (AI) workloads, including conversational AI and code generation. 
    We introduce a modeling approach that approximates real-world LLM workflows through a binning strategy for input-output token distributions and batch size variations. Our empirical analysis spans software frameworks, decoding strategies, GPU architectures, online and offline serving settings, and model parallelism configurations. 
    We show that the effectiveness of inference optimizations is \textit{highly sensitive to workload geometry, software stack, and hardware accelerators}, demonstrating that naive energy estimates based on FLOPs or theoretical GPU utilization significantly underestimate real-world energy consumption.
    Our findings reveal that the proper application of relevant inference efficiency optimizations can reduce total energy use by up to \textbf{73\%} from unoptimized baselines. These insights provide a foundation for sustainable LLM deployment and inform energy-efficient design strategies for future AI infrastructure.

\end{abstract}

\section{Introduction}
\definecolor{theoreticalgreen}{RGB}{0,153,0}
\definecolor{pytorchpurple}{RGB}{128,0,128}
\definecolor{vllmorange}{RGB}{255,153,51}

\label{sec:introduction}

Improvements in task performance by large language models (LLMs) have prompted large-scale investments in computing hardware and energy infrastructure to support the development and deployment of LLM and related machine learning models \citep{nytmeta2025, msft2025, google2025}. However, the growing prevalence of LLMs yields commensurate increases in the energy demand, water use, and carbon emissions associated with their development and deployment \citep{morrison2025holistically, li_making_2025, strubell2020energy,luccioni2024power}. Primarily motivated by the increased demands from LLM and AI workloads, projections estimate that that data centers consume between 9.1\% and 11.7\% of the total US energy demand by 2030 \citep{aljbour2024powering, shehabi20242024, mckinsey2024energy}. However, such projections of energy use primarily rely upon sector-wide estimates of demand or substantial simplifications of the of the energy demands of individual models.

\begin{figure}
    \centering
    \vspace{-2em}
    \includegraphics[width=0.85\linewidth]{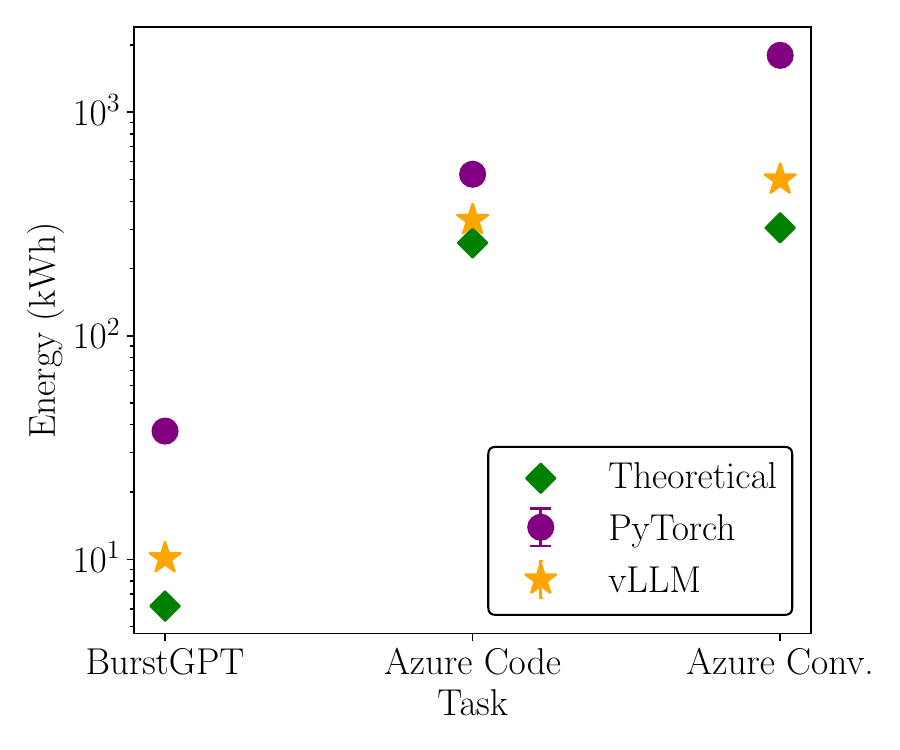}
    \vspace{-1em}
    \caption{
    Proper application of efficiency methods with optimized vLLM ({\textcolor{vllmorange}{orange}}) approaches the ideal energy consumption ({\textcolor{theoreticalgreen}{green}}) as compared with an unoptimized baseline PyTorch ({\textcolor{pytorchpurple}{purple}}) implementation.
    }
    \vspace{-2em}
    \label{fig:figure-1}
\end{figure}
 
In order to develop effective energy policy for this growing demand, it is necessary to characterize the underlying computational workloads of development (i.e. model training) and deployment (i.e. inference). In particular, the cost and efficiency of inference is especially crucial due to the scale and increased frequency at which models are served for repeated use. Concretely, Meta reports that inference workloads constitute up to $70\%$ of their AI power consumption \citep{wu2022sustainable} while Google attributes $60\%$ of their ML energy \citep{patterson2022carbon} and between 80 to 90\% of ML AWS cloud computing demand \citep{barr2019amazon, leopold2019aws}.

To address the problem of inference efficiency, the NLP and machine learning research communities have developed various optimizations spanning: algorithms, software frameworks, and hardware accelerators. Such optimizations have primarily targeted improvements in model speed (e.g. latency and throughput; \cite{leviathan2023fast,kwon2023efficient}).
% or reductions in the computational requirements of models (e.g. FLOPs; ).
Moreover, these methods are frequently assessed in constrained settings or on simplified datasets that fail to capture the broad diversity of real-world tasks. These tasks range from traditional NLP applications like sequence tagging and summarization to more computationally demanding workloads such as synthetic data generation and chain-of-thought reasoning. \textit{There remains a critical gap in understanding of the energy costs of language model inference, especially when efficiency interventions are applied jointly in real-world settings.}

In this work, we examine the energy costs of LLM inference and present a comprehensive analysis of the impact of: \textit{data dimensionality, decoding strategies, serving frameworks, compilation techniques, GPU hardware platforms, model parallelism, and architectural variants} on total energy use during inference.  Based on our energy profiling across these optimizations, we approximate offline inference with LLMs based on real-world workload with variable sequence lengths and batching, considering both an upper bound of naive unoptimized inference and a lower bound of theoretical optimized inference. 
Our analysis reveals that while idealized estimations of hardware utilization substantially underestimate the energy use of language model inference, proper application of inference efficiency optimizations can substantially \textbf{reduce the energy requirements of inference by up to 73\% from unoptimized baselines with vanilla PyTorch and Huggingface Transformers} and to within 26.6\% of theoretical ideal performance on simulated offline workloads (see Table \ref{tab:energy_percentage_diff}).

\section{Methodology}
\label{sec:methods}
In the following section, we describe our experimental setup for evaluating inference efficiency.

\paragraph{Model Architectures.}
We focus our experiments on language models ranging from 1B to 32B parameters, primarily evaluating \wrapt{Llama-3.1-8B-Base} and \wrapt{Llama-3.1-8B-Instruct} models as representative decoder-only transformers \cite{dubey2024llama}. To investigate effects of scaling model architecture, we include the \wrapt{Qwen-1.5-32B} model \cite{bai2023qwen}. For architectural comparisons, we analyze the sparse \wrapt{OLMoE} mixture-of-expert (MoE) model alongside its dense counterparts -- the 1B and 7B \wrapt{OLMo} architectures -- which maintain comparable active and total parameter counts, respectively \cite{muennighoff2024olmoe, groeneveld2024olmo}.

\paragraph{Data Dimensionality}

We investigate the impact of data dimensionality across three key dimensions: input sequence lengths, output generation lengths, and batch sizes.

Inference with large language models is commonly decomposed into two stages: prefilling and token generation, each with a different energy profile \cite{patel2024characterizing}. The prefill stage processes prompts in parallel and is typically compute-bound, achieving high GPU utilization. In contrast, the autoregressive decoding stage is typically memory-bound and leads to GPU under-utilization. These bottlenecks and their resulting energy profiles shift with input and output lengths.

To address GPU under-utilization during generation, serving systems employ batching strategies. However, the effectiveness of batching varies with input-output characteristics \citep{agrawal2024tamingsarathi, li2024sprout}. Long input sequences limit maximum batch sizes due to memory constraints, while variable output lengths can lead to inefficient batch utilization as some sequences complete before others.

Our analysis spans batch sizes from 1 (single-example inference) to task-dependent maximums (up to 1024), ensuring coverage of a broad range of maximally realistic settings.

We ground analysis in NLP workloads spanning text classification, summarization, translation, and open-ended text generation. Different tasks exhibit different data dimensionalities: classification involves minimal generation (often a single token), summarization pairs long contexts with medium-length outputs, and translation typically assumes balanced input-output lengths. Input length statistics in considered datasets are shown in Table \ref{tab:nlp-token-stats}.

In a controlled sweep, we explore scenarios with up to 32k input tokens and 4k output tokens, varying sequence lengths by powers of two. We fix generation to 64 or 8 tokens when varying context lengths, and assume 512 or 64 token input context when varying output lengths. Input context length is enforced via truncation of longer sequences from PG19 \citep{raecompressive2019}.

\begin{figure*}[t!]
        \centering
        \includegraphics[width=\textwidth]{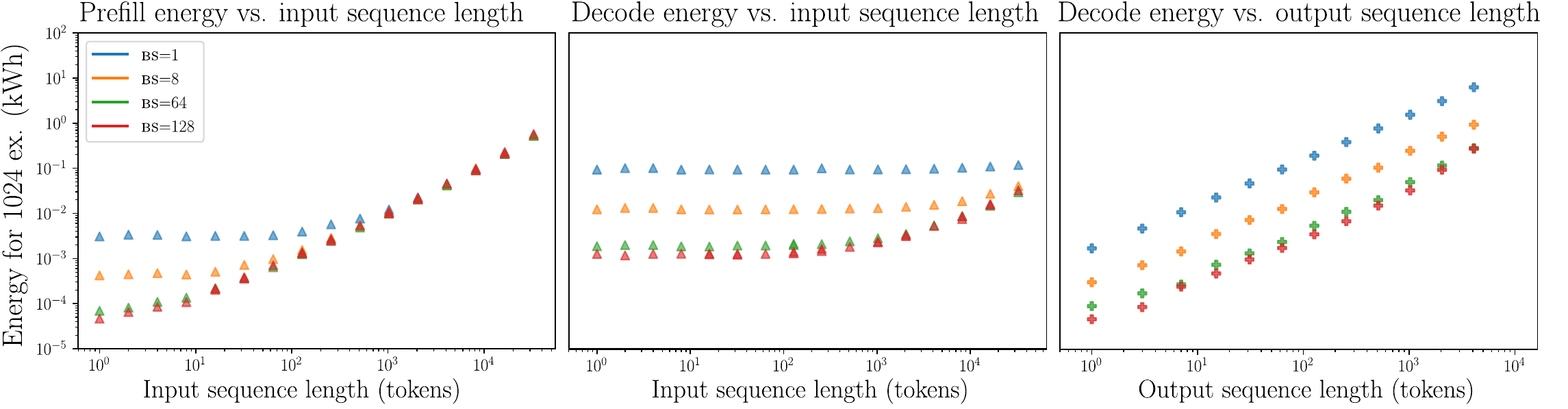}
        \caption{
            Controlled sweeps of input and output sequence lengths on A6000 GPUs, on vLLM backend, described in \S\ref{subsec:seqlen_results}. We decompose inference costs into prefill and decode energy. At small batch sizes and input sequence lengths, energy intensity of a workload scales sub-linearly with increasing sequence length input sequence lengths. Decoding is more energy intensive per token than prefill, but energy intensity begins scaling linearly even for short generations and small batch sizes with the vLLM framework.
            }
            \vspace{-0.5em}
            % this belongs to section 3 -- just want it on page 3
        \label{fig:inout-seqlen}
    \end{figure*}

\paragraph{Decoding Strategies.} Different decoding strategies used for generation have different computational profiles and can have a substantial impact on the generation efficiency \cite{kwon2023efficient}. In order to study the impact of sampling methods and auto-regressive decoding strategies, we investigate \textit{greedy decoding, beam search decoding, temperature sampling, top-$p$ decoding} affect the energy requirements and end-to-end latency \cite{Holtzman2020The}.

In addition to auto-regressive decoding, we study the impact of speculative decoding. \textit{Speculative decoding} is commonly used as a latency minimization inference optimization \cite{kwon2023efficient}. In speculative decoding, a lightweight draft model is used to predict multiple tokens ($\gamma$) which are then verified by the target model in parallel \cite{leviathan2023fast, chen2023accelerating}. Speculative decoding provides latency improvement by better utilizing GPUs over autoregressive decoding.

In our experiments, we use the following target-draft model pairs with a look-ahead value $\gamma=4$ across various batch sizes: \wrapt{DeepSeek-R1-Distill-Qwen-32B} with \wrapt{mobiuslabsgmbh/DeepSeek-R1-ReDistill-Qwen-1.5B-v1.1} \cite{guo2025deepseek, yang2024qwen2}; \wrapt{Llama-3.1-8B-Base} with \wrapt{Llama-3.2-1B} \cite{dubey2024llama}.% \todo[size=\small]{footnotes maybe?}
%% https://ai.meta.com/blog/llama-3-2-connect-2024-vision-edge-mobile-devices/

\paragraph{Software Optimizations.}
Choice in the software frameworks used for inference significantly impacts both latency and energy efficiency through optimized kernel implementations and computational graph management \cite{georgiou2022green, fernandez2023framework}. We evaluate two widely-adopted libraries used in LLM inference: native PyTorch with HuggingFace transformers~\cite{wolf2020transformers}, and vLLM, an optimized framework for LLM inference that achieves improved compute and memory utilization \cite{paszke2019pytorch, kwon2023efficient}; experiments are conducted in \texttt{bfloat16} precision.

Within these frameworks, we compare with a native PyTorch baselines with Just-in-Time compilation via TorchInductor (i.e. \texttt{torch.compile}) and CUDA Graphs kernel serialization. 
Furthermore, for vLLM, we evaluate continuous batching which efficiently handles variable output lengths in batch processing by overlaying sequences \cite{yu2022orca}.

\paragraph{Hardware Platforms.}
Our experiments are conducted using an on-premise heterogeneous server with multiple GPU types and node configurations.
Specifically, we conduct experiments on multiple generations of consumer workstation and datacenter GPU accelerators from the Ampere (A6000, A100 80GB PCIe), and Ada Lovelace (A6000 Ada) microarchitecture.

All experiments run on 8-GPU nodes with standardized node- and job-level CPU and RAM configurations for each GPU type. For multi-GPU experiments, we utilize up to 4 GPUs simultaneously, investigating tensor parallel inference with group sizes of 2 and 4 devices. \footnote{This configuration leaves 4-7 GPUs available for other users. While the Slurm scheduler does not enforce complete isolation in network, memory, and CPU infrastructure across jobs, concurrent workloads in practice were not CPU- or memory-intensive enough to impact ours significantly -- for example, in the vast majority of cases (98\%), an ambient measurement of the RAM utilization in a node our jobs were running on was less than $20\%$ of the total available}. We examine both standard and speculative decoding approaches using the \wrapt{Llama-3.1-8B} and \wrapt{Qwen-32B} models. Additional details on computing hardware are provided in Appendix \ref{appx:hardware}.

\paragraph{Performance Measures.}
We evaluate the efficiency of inference by measuring the latency, throughput, GPU energy, and GPU power required for the inference of 1,024 examples
\footnote{For experiments with batch sizes larger than 256, metrics are computed over 4096 examples and then normalized.}. Total energy use and GPU power metrics are measured using Nvidia Management Library (NVML) via the \texttt{CodeCarbon} library \cite{benoit_courty_2024_11171501}.
Prior to evaluation, we conduct a warmup on up to 20 batches to allow for memory allocation, required CUDA graph capture, and JiT compilation \footnote{Due to size, warmup is limited to 4 batches for inference with the \texttt{Qwen-32B}.}. Results are reported as the mean values energy use, latency, or power usage of three runs.

    \begin{figure}
        \centering
        \includegraphics[width=0.85\linewidth]{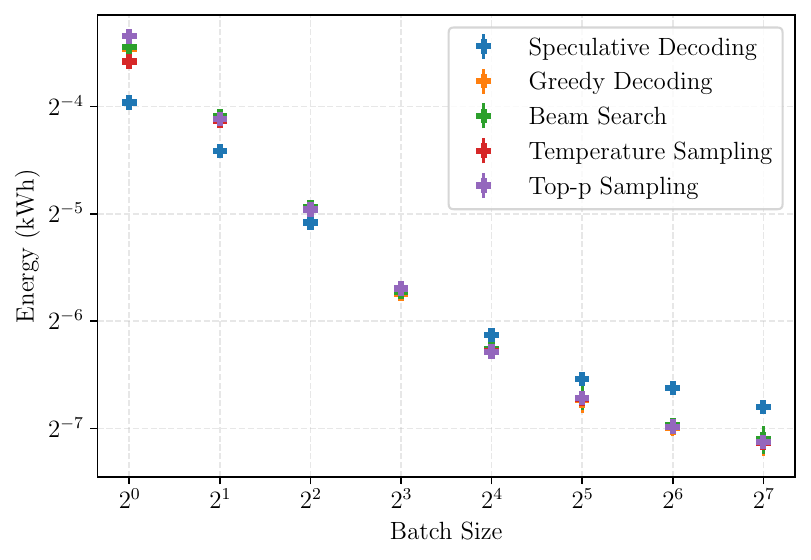}
        \caption{
            At small batch sizes, speculative decoding provides reduced latency and energy savings. At larger batch size speculative decoding increases energy.
        }
        \vspace{-1em}
        \label{fig:decoding-methods}
        % \vspace{-1.5em}
    \end{figure}
    
\section{Results}
\label{sec:results}
In the following section, we examine the effects of variations of data dimensionality, model architecture, decoding strategies, and software optimizations on inference energy use.

\subsection{Effects of Dataset and Sequence Length}
\label{subsec:seqlen_results}

We present results from our controlled sweep of sequence lengths and batch sizes in Figure~\ref{fig:inout-seqlen}. Prefill costs increase as a function of input sequence length, \textit{at the same rate} regardless of batch sizes when scaling sequences larger than 128 tokens.
At shorter sequence lengths and smaller batch sizes, the energy costs of prefill are constant regardless of the computational workload due to significant undersaturation of the accelerator.
Although we fix output generation tokens to 64, we verify that at this convergence in rate of energy intensity increase occurs at the same point when instead fixing generation length to 8 tokens; see Figure~\ref{fig:inout-seqlen-compare} in Appendix~\ref{appx:seqlen}. 

In Figure \ref{fig:inout-seqlen}, the energy intensity of the decode likewise scales with input context length only at larger input sequence lengths.

However, the energy intensity of decoding scales linearly with sequence length regardless of sequence length or batch sizes due to the autoregressive, sequential nature of decoding.

Generally, decoding energy dominates the overall workload in all settings but those with the shortest generation lengths, such as those seen in classification workloads and short form summarization. Note the log-log scale and the parallel linear trends, where the differences in intercepts are proportionate with the differences in batch size 
\footnote{See Fig~\ref{fig:inout-seqlen-pytorch} in Appendix~\ref{appx:seqlen} for additional results on vanilla PyTorch backend, and Figure~\ref{fig:tasks-vllm} for comparison with real energy intensity measurements for a sample of classical NLP tasks}.
In the following sections, we discuss a variety of algorithmic and software interventions that are appropriate for different types of workload geometries.

    \begin{figure}
        % \vspace{-23em}
        \centering
        \includegraphics[width=0.85\linewidth]{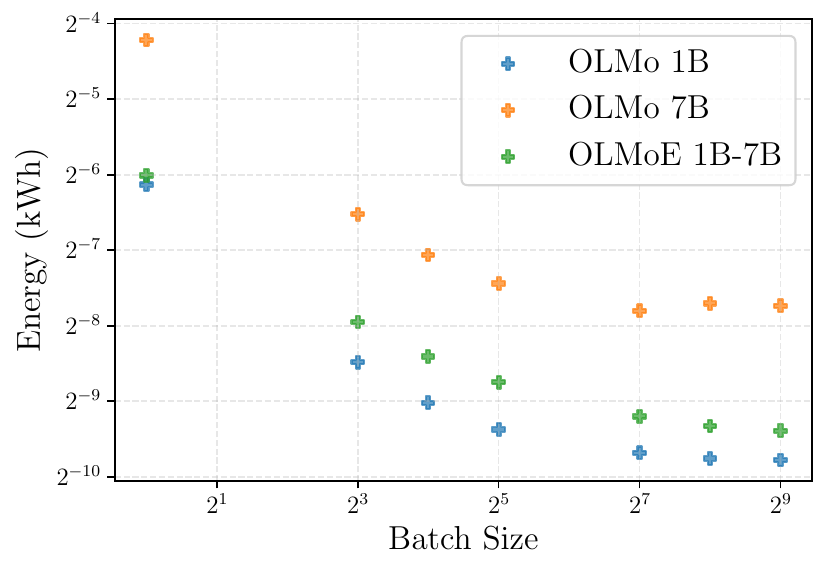}
        \caption{
            Mixture-of-Experts LLMs require more energy than dense models with comparable active parameters; differences are pronounced at larger batch sizes. 
        }
        \label{fig:moe}
        \vspace{-1em}
    \end{figure}

\begin{figure*}
        \centering
        \begin{subfigure}{0.3\textwidth}
            \centering
            \includegraphics[width=\linewidth]{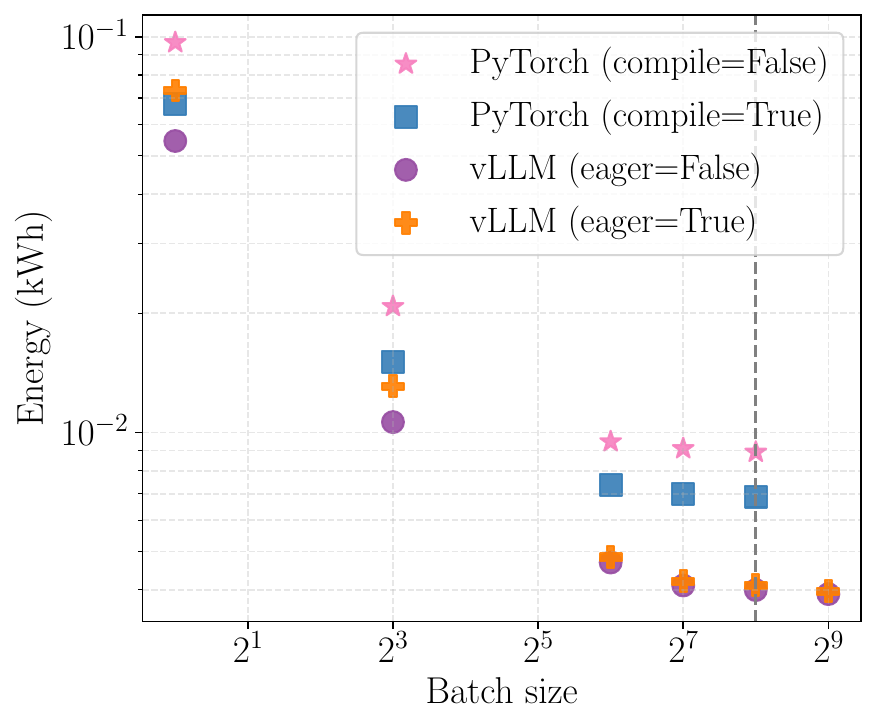}
            \caption{A100 80GB PCIe}
        \end{subfigure}
        \hfill
        \begin{subfigure}{0.3\textwidth}
            \centering
            \includegraphics[width=\linewidth]{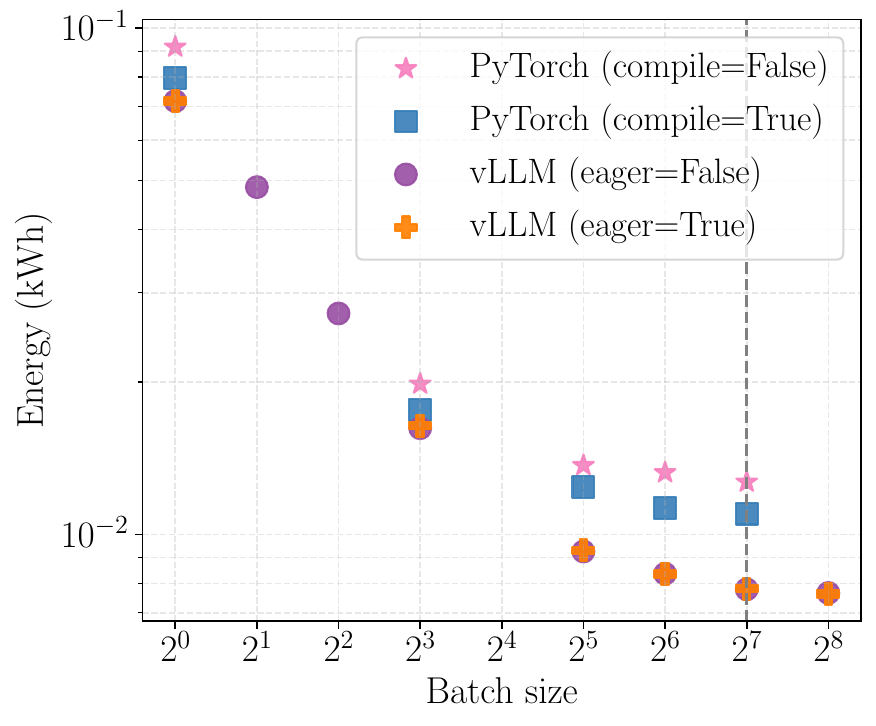}
            \caption{A6000 Ada}
        \end{subfigure}
        \hfill
        \begin{subfigure}{0.3\textwidth}
            \centering
            \includegraphics[width=\linewidth]{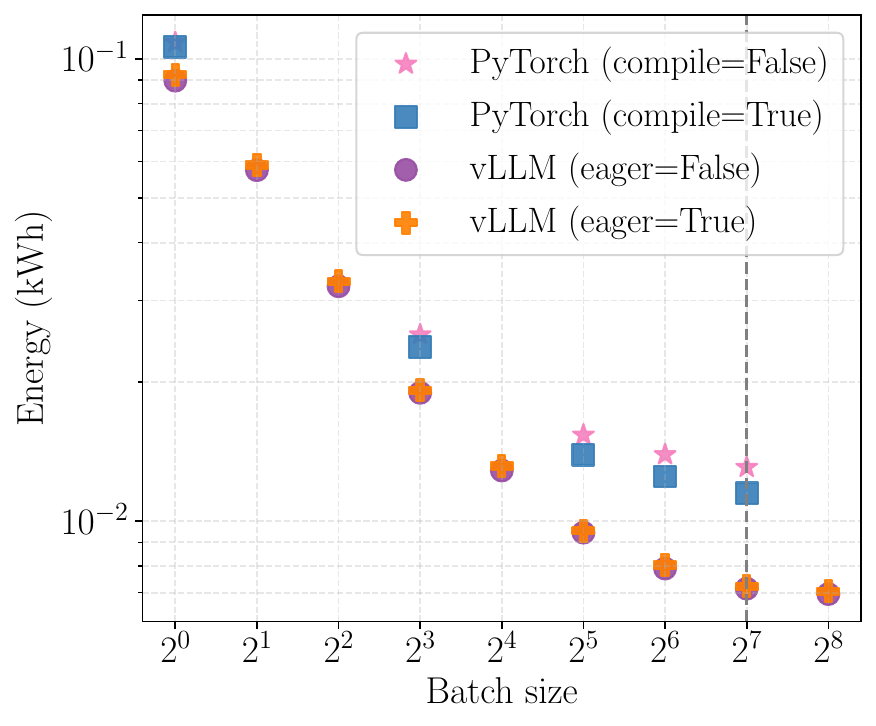}
            \caption{A6000}
        \end{subfigure}
        \caption{
            {Energy consumption comparison across different GPUs for inference with PyTorch and vLLM backends of 1024 samples for 64 output tokens.} For each GPU, we compare PyTorch with and without compilation, and vLLM with and without CUDA Graph serialization. The line in black represents the maximum allowable batch size for PyTorch. Relative savings are most apparent in the low batch size regime and that vLLM due to its optimizations can serve a larger batch size.}
        \label{fig:software-opt}
        \vspace{-1em}
    \end{figure*}

\subsection{Effects of Algorithmic Optimizations}
\paragraph{Speculative Decoding Only Reduces Energy at Low Batch Sizes.}
\label{subsec:decoding_results}

Speculative decoding is commonly used to achieve inference speedups in low-batch inference in which autoregressive decoding fails to achieve high GPU VRAM utilization. However, for large batch sizes where GPU is already saturated, draft model speculation and excess verifications introduce additional overhead. In the large batch case, for short to medium contexts, LLM inference is typically compute bound, making speculative decoding slower than autoregressive decoding with the target model \cite{chen2024magicdec, liu2024optimizing}.

Compared to variations in energy use from alternate decoding strategies and sampling methods, speculative decoding has the greatest effect on the energy use and latency of language model inference. At smaller batch sizes ($\leq16$) speculative decoding is effective in reducing the total energy cost of inference with up to $+29.14\%$ compared to  single-example inference (Figure \ref{fig:decoding-methods}). However, autoregressive decoding methods are more efficient at larger batch sizes, with speculative decoding requiring $25.65\%$ more energy when performing inference at a batch size of 128.

\vspace{-.25em }
\paragraph{Mixture of Experts Incurs Higher Inference Energy Costs.} Sparse mixture-of-experts are often utilized as an alternative architecture due to their increased sample efficiency during training and increased performance relative to dense neural networks with the same number of active parameters. Although both dense \wrapt{OLMo-1B} and the \wrapt{OLMoE 1B-7B} mixture-of-experts models use substantially less energy than the dense \wrapt{OLMo-7B} model, the OLMoE architecture utilizes up to $\bf{54.24\%}$ more energy than the base OLMo 1B model, despite having a similar number of active parameters. 

We identify that the increased energy and latency of MoE’s can be attributed to the fused kernel used in the expert layers which is substantially slower than the corresponding GEMM operation in linear layers in the dense model; 19.70\% slower at batch size 1 and 63\% slower at batch size 8. Notably, we observe that the additional routing operations in the MoE model introduce minimal latency; and that the increased overhead of more CUDA graph and kernel launch operations are largely mitigated through kernel serialization and graph compilation optimizations (i.e. vLLM with CUDA Graphs).

\subsection{Effects of Software Optimizations}
\label{subsec:frameworks_results}

    \paragraph{PagedAttention with vLLM Improves Efficiency.}
    Compared to native PyTorch, the vLLM inference serving engine improves both the throughput and the energy efficiency.
    The vLLM framework uses PagedAttention to implement non-contiguous KV cache blocks which reduces memory fragmentation and allocation of redundant memory in the case of sparse sequences \cite{kwon2023efficient}.

    These optimizations allow for improved memory efficiency and the vLLM framework to support larger batch sizes on fixed memory GPUs.

    \paragraph{Compilation and Kernel Serialization Improves Efficiency.}
    The graph compilation and kernel serialization increase hardware utilization by removing redundant operations in the computational graph and reducing the kernel launch overhead \citep{fernandez2023framework}, respectively.
    We observe that both \texttt{torch.compile} and CUDA graph serialization (\texttt{eager=False}) improve throughput at no additional energy cost in Figure \ref{fig:software-opt}. However, we note that the benefits of CUDA graphs are more apparent at lower batch sizes, as the relative cost of kernel launch is larger for smaller computational workloads.

    \begin{figure}
        \centering
        \begin{subfigure}{0.425\textwidth}
            \centering
            \includegraphics[width=\linewidth]{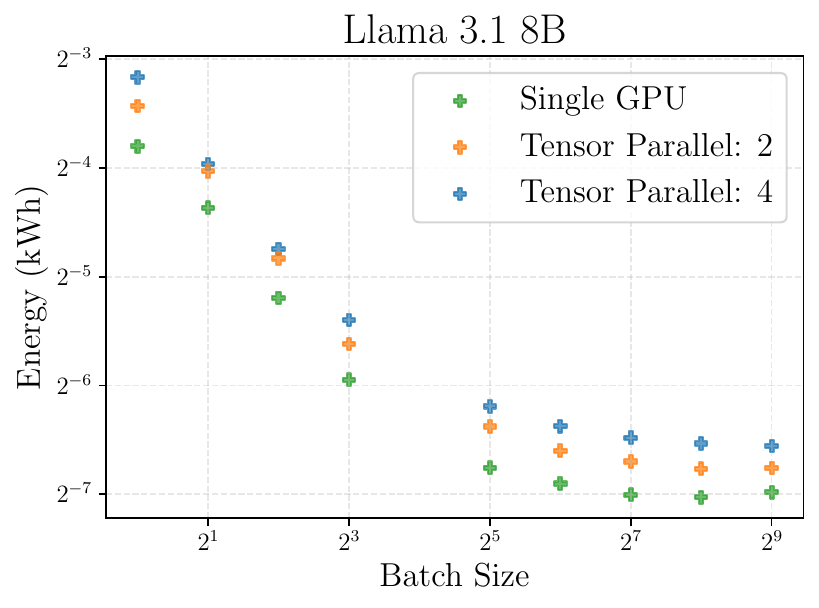}
        \end{subfigure}
        \hfill
        \begin{subfigure}{0.425\textwidth}
            \centering
            \includegraphics[width=\linewidth]{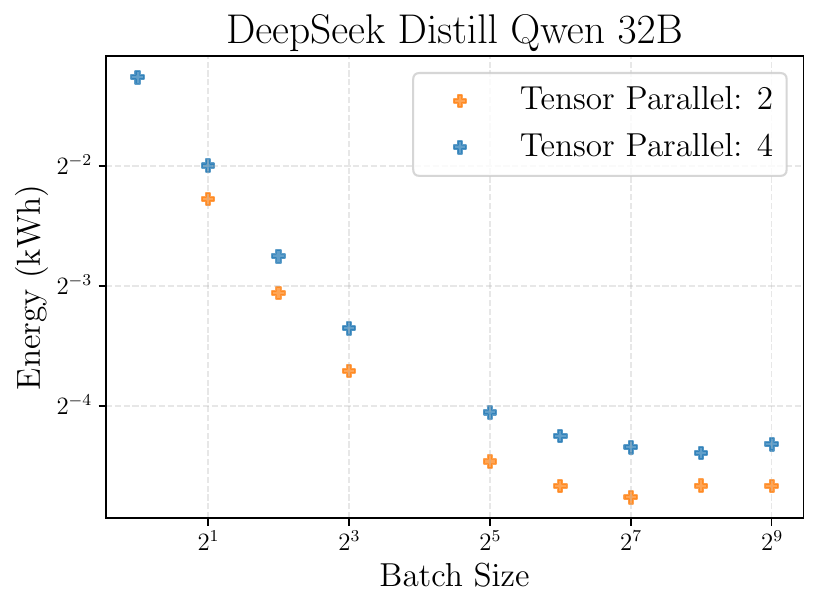}
            % \caption{A6000}
        \end{subfigure}
        \vspace{-0.5em}
        \caption{Energy Use of Llama-3.1 8B and Qwen 32B with varying degrees of Tensor Parallelism.}
        \label{fig:multigpu}
        \vspace{-1em}
    \end{figure}

   \paragraph{Continuous Batching Reduces Energy Use.}
    LLM inference is inherently autoregressive, requiring many sequential operations. Static batching maintains a fixed batch size throughout inference, which leads to GPU under-utilization when generation lengths vary and idle compute accumulates after early terminations. \textit{Continuous batching} mitigates this by dynamically replacing completed requests with new ones, improving GPU utilization and reducing idle time \cite{yu2022orca}. This approach is particularly effective when generation lengths have high variance, yielding significant speedups at larger batch sizes.

    We observe that at smaller batch sizes the overhead of online scheduling outweighs its benefits but at larger batch sizes, online serving with continuous batching requires less energy; details in Appendix \ref{appx:software}. We note that the numbers under-represent the impact of continuous batching given the samples are drawn from the same dataset, thereby reducing the variance in input and output lengths.

\subsection{Effects of Hardware Design Choices}
\label{subsec:hardware_results}

\paragraph{Multi-GPU Tensor Parallelism Reduces Latency for Increased Power Use}

    Model parallelism techniques such as tensor and pipeline parallelism are frequently used to alleviate the memory pressure of large sets of model parameters and batch sizes, as well as to leverage multiple hardware accelerators in order to speed up workload execution \cite{narayanan2021efficient}. Additionally, for fixed workloads, tensor parallelism reduces both the per-device computational intensity and per-device power utilization as the workload is sharded across accelerator.
    % \jared{Reference the proper strong and weak scaling argument here}
    However, the speedups from additional accelerators are insufficient to offset the energy cost of utilizing more devices  (i.e. utilizing twice the GPUs fails to yield a two-fold speedup).

    In Figure \ref{fig:multigpu}, we observe that utilizing tensor parallelism to scale from inference with a single GPU to four GPUs reduces latency and per-device power utilization for the Llama-3.1 8B model. However, increasing parallelism yields higher total energy use due to the larger number of accelerators. Concretely, parallelizing a fixed workload over two and four GPUs decreases latency by 40.16\% and 61.34\% but increases total energy use by 29.3\% and 55.23\% at single batch inference due to the introduction of additional devices.

    \paragraph{Effects of Hardware Speed}
    The effectiveness of optimization techniques varies significantly across hardware platforms, with faster accelerators showing greater benefits from optimizations that target computational efficiency. Our results demonstrate that graph compilation, kernel serialization, and speculative decoding achieve their maximum impact on the A100 GPU.

    Specifically, PyTorch compilation yields a 29.90\% improvement on the A100, which drops to 13.28\% on the RTX 6000 Ada and further to 1.96\% on the A6000. Similarly, vLLM's eager mode optimization shows a 25.47\% improvement on the A100 versus 2.97\% on the A6000. This pattern suggests that as hardware computational capabilities increase, the relative impact of software optimizations targeting kernel efficiency becomes more pronounced.

\section{The Impact of Optimizations on Inference Energy Use}
\label{sec:modeling-energy}

\begin{figure*}
    \centering
    \begin{subfigure}{0.4\textwidth}
        \centering
        \includegraphics[width=\linewidth]{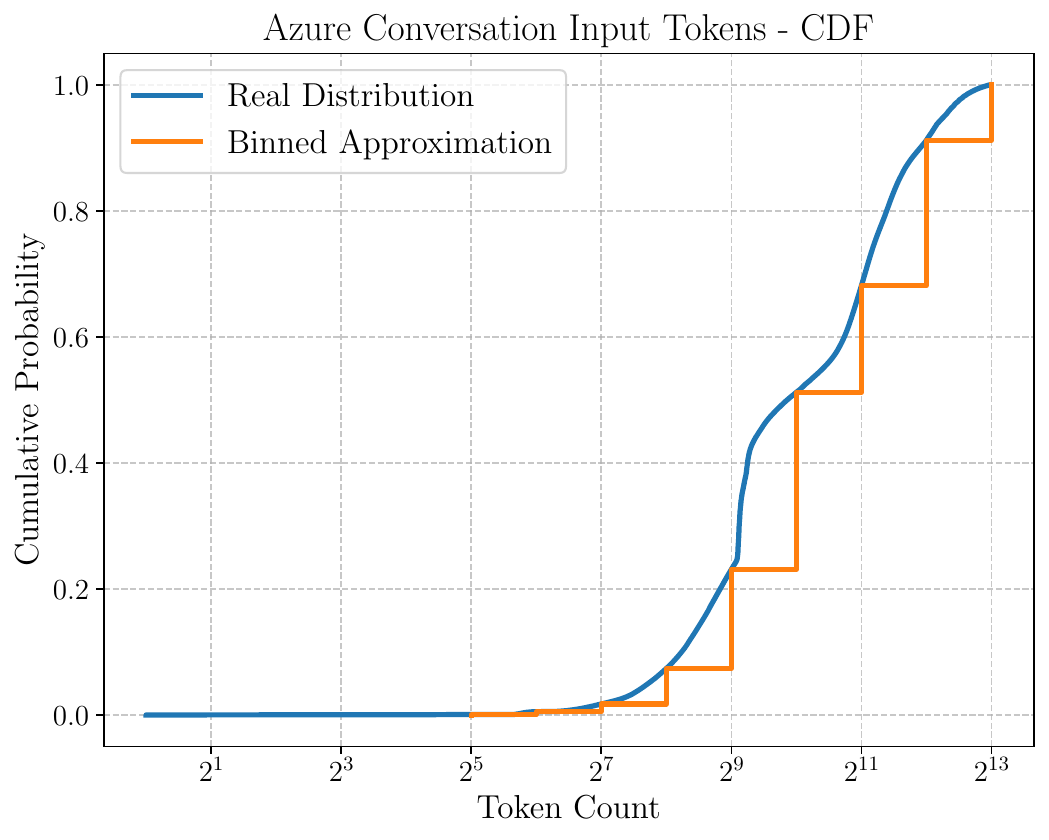}
    \end{subfigure}
    \hfill
    \begin{subfigure}{0.4\textwidth}
        \centering
        \includegraphics[width=\linewidth]{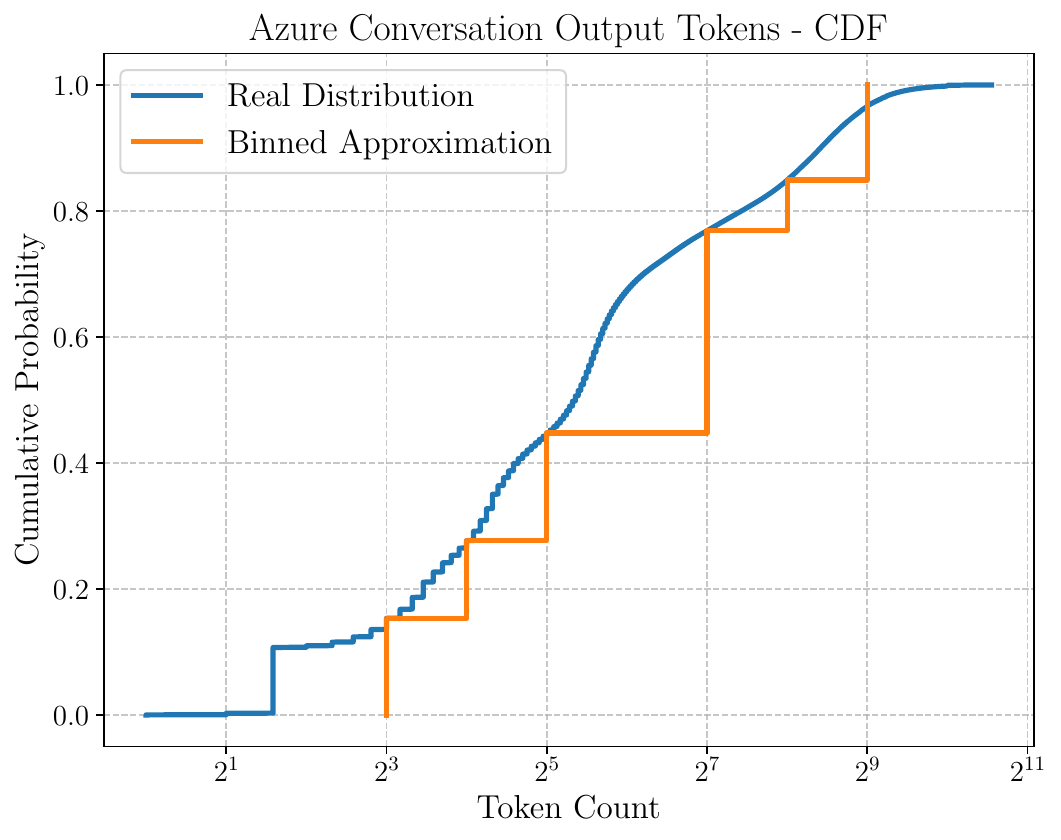}
    \end{subfigure}
    \caption{Comparison of the real token length distributions (blue) with the binned approximation (orange) for Azure conversation input (left) and output (right) token lengths. The CDF plots illustrate how our binning strategy approximates the empirical distribution while ensuring computational efficiency for energy estimation.}
    \label{fig:modeling-azure-conv}
    \vspace{-1em}
\end{figure*}

In this section, we outline our approach to modeling the energy consumption of an LLM under both synthetic and realistic workload distributions. We leverage classical NLP tasks and datasets of inference requests to estimate energy usage across different execution environments, including PyTorch-native and vLLM backends with software optimizations on a single A6000 GPU.

\subsection{Modeling Energy Requirements Using Offline Serving}
\label{subsec:model-description}
We consider the energy required to process a  dataset \( \mathcal{D} = \{R_1, R_2, \dots, R_N\} \) in an offline setting in which all requests can be batch processed freely, and where each request \( R_k \) consists of a tuple \( (i_k, o_k) \), representing the input token length \( i_k \) and the output generation length \( o_k \):
\begin{equation*}
    R_k = (i_k, o_k), \quad \forall k \in \{1, \dots, N\}.
\end{equation*}
Since \( i_k \) and \( o_k \) vary significantly across requests, we utilize dataset statistics—including the median and 99th percentile of input and output lengths (discussed in \S\ref{subsec:model-evaluation}) to inform our binning strategy.

\paragraph{Binning Strategy.}
To effectively handle the broad range of \( (i_k, o_k) \) values, we define discrete bin sets for input and output lengths:
\begin{align*}
    I_{\text{bins}} &= \{2^m \mid m \in \mathbb{N}, 4 \leq m \leq 13 \} \\
    &= \{32, 128, 256, 512, 1024, 2048, 4096, 8192\}, \\
    O_{\text{bins}} &= \{2^n \mid n \in \mathbb{N}, 3 \leq m \leq 9 \} \\
    &= \{8, 16, 32, 64, 128, 256, 512\}.
\end{align*}
These bin choices ensure sufficient coverage across realistic request distributions. % while aligning with CUDA kernel optimization strategies for efficient inference execution. 
Notably, we exclude extremely long input requests (\(>8k\) tokens) and generation outputs beyond 512 tokens.
% \vash{check the cuda kernel stuff and add citation}

\paragraph{Mapping Requests to Bins.}
Given a request \( R = (i, o) \), we map it to the closest ceiling bin:
\begin{align*}
    I^* &= \min\{I \in I_{\text{bins}} \mid I \geq i\}, \\
    O^* &= \min\{O \in O_{\text{bins}} \mid O \geq o\}.
\end{align*}

%add what B is (-Vash max allowable)
We group requests within the same \( (I^*, O^*) \) bin into batches of size \( B(I^*, O^*) \), the maximum allowable batch size for the given hardware and backend configuration. Each batch processes \( B(I^*, O^*) \) requests in parallel, allowing for more efficient energy utilization, which is more representative of real-world inference setups.

Given our hardware configuration and backend, we collect the estimates of  $E_{\text{batch}}(I^*, O^*)$, which corresponds to the energy used to serve a request of batch size $B$ with input prompts of length $I*$ and output lengths $O*$.

We collect real energy measurements \( \bf{E_{\text{batch}}^{\text{real}}(I^*, O^*)} \), representing the observed energy usage when processing a full batch of size \( B(I^*, O^*) \) with input lengths \( I^* \) and output lengths \( O^* \).  Thus, the total estimated energy consumption across the workload to serve $N$ requests that fall in the bin is given by:
\begin{equation*}
    \widehat{E}_{\text{total}} = \sum_{(I^*, O^*)} \left( \frac{N^{\text{real}}(I^*, O^*)}{B(I^*, O^*)} \right) E_{\text{batch}}^{\text{real}}(I^*, O^*),
\end{equation*}
where \( N^{\text{real}}(I^*, O^*) \) is the total number of observed requests mapped to bin \( (I^*, O^*) \), and \( \frac{N^{\text{real}}(I^*, O^*)}{B(I^*, O^*)} \) represents the number of batches required to process them.

\subsection{Idealized Baseline}
As a naive baseline, we estimate an upper bound of the energy efficiency of these workloads with a baseline derived from the manufacturer-rated hardware speeds ($FLOPS_{HW}$), power draw (TDP) ,and floating point operations (FLOPs) required for inference $FLOPs$
\footnote{Based on the Nvidia datasheet for the RTX A6000 GPU, we utilize consider $FLOPS_{HW}$ of 309.7 TFLOPS and a 300W TDP power draw; and estimate theoretical inference FLOPs with the DeepSpeed profiler \cite{rasley2020deepspeed}.}. This approximation assumes hardware is being utilized as maximum efficiency both in through idealized floating point operation throughput and maximum power draw. 

\begin{align*}
    \widehat{E}_{\text{Optimal}} &= 
    \left(\frac{\text{TDP}}{FLOPS_{HW}}\right) \\
    & \times  \sum_{(I^*, O^*)} N^{real}(I^*, O^*) 
    \times FLOPs(I^*, O^*) 
\end{align*}

\subsection{Evaluations}
\label{subsec:model-evaluation}

\begin{table}[ht]
    \centering
    \footnotesize
    \begin{tabular}{lccc}
    \toprule
        \textbf{Dataset} & \textbf{Mean} $\pm$ \textbf{Std} & \textbf{Median} & \textbf{99th} \\
    \midrule
        BurstGPT          & 256.80  $\pm$ 242.27   & 215  & 1038     \\
        Azure Chat & 1631.58 $\pm$ 1529.64 & 928  & 6683     \\
        Azure Code & 2511.28 $\pm$ 2133.54 & 1930 & 7685     \\
        % Azure Chat (Day)  & 1154.70 $\pm$ 1108.82 & 1020 & 4142 \\
        % Azure Code (Day)  & 2047.85 $\pm$ 1973.88 & 1469 & 7436\\
    \bottomrule
    \end{tabular}
    \caption{
        Input Sequence Length Statistics Across Real-World LLM Workloads 
    }
    \label{tab:llm-token-stats-in}
\end{table}

\begin{table}[h]
    \centering
    \footnotesize
    \begin{tabular}{lccc}
    \toprule
        \textbf{Dataset} & \textbf{Mean} $\pm$ \textbf{Std} & \textbf{Median} & \textbf{99th} \\
    \midrule
        BurstGPT    &  35.10 $\pm$ 108.59   & 7     & 478     \\
        Azure Chat  & 105.51 $\pm$ 158.25   & 41    & 694     \\
        Azure Code  &  22.69 $\pm$ 74.78    & 8     & 271     \\
        % Azure Chat (Day)  & 211.13 $\pm$ 162.87 & 129 & 601\\
        % Azure Code (Day)  & 27.88  $\pm$ 59.86 & 13 & 251.46 \\
    \bottomrule
    \end{tabular}
    \caption{
        Output Sequence Length Statistics Across Real-World LLM Workloads
    }
    \vspace{-1em}
    \label{tab:llm-token-stats-out}
\end{table}

% \begin{table}[h]
% \centering
% \small
% \begin{tabular}{lcccc}
% \toprule
%     \textbf{Task} & \textbf{Mean $\pm$ Std} & \textbf{95th} \\
% \midrule
%     Translation & 49.96 $\pm$ 39.39 & 109.70 \\
%     % wmt14-fr-en-sample & en & 32.77 & 25.19 & 547 & 72.85 \\
%     Generation & 136.89 $\pm$ 93.13 &  310.70 \\
%     Classification & 292.48 $\pm$ 239.94 &  736.20 \\
%     Summarization & 838.49 $\pm$ 400.70 & 1647.85 \\
%     % cnn\_dailymail & highlights & 64.49 & 24.83 & 199 & 111.85 \\
% \bottomrule
% \end{tabular}
% \caption{
%     Input Length Statistics Across NLP Datasets 
%     % \jared{This can be appendixed; What are the medians and output lengths?}
% }
% \label{tab:nlp-token-stats}
% \end{table}

\begin{table}[h]
\centering
\small
\begin{tabular}{lccc}
\toprule
\textbf{Task} & \textbf{Mean $\pm$ Std} & \textbf{Max} & \textbf{Output} \\
\midrule
\textbf{Translation} & 49.96 $\pm$ 39.39 & 550 & 64 \\
\textbf{Generation}       & 136.89 $\pm$ 93.13 & 547  & 64 \\
\textbf{Classification}   & 292.48 $\pm$ 239.94 & 3112  & 1  \\
\textbf{Summarization} & 838.49 $\pm$ 400.70 & 2386 & 64 \\
\bottomrule
\end{tabular}
\caption{
    Tokenized Input and Output Length Statistics Across NLP Tasks used for Energy Benchmarking
}
\vspace{-0.5em}
\label{tab:nlp-token-stats}
\end{table}

We examine a suite of classical NLP tasks and LLM inference workloads, each characterized by a range of different input context and output generation sequences; with dataset statistics provided in Tables \ref{tab:nlp-token-stats}, \ref{tab:llm-token-stats-in}, \ref{tab:llm-token-stats-out}. We simulate a large-scale offline processing setting on the RTX A6000 GPUs, in which examples are binned by sequence lengths (as described in \S \ref{sec:modeling-energy} and processed in parallel in the largest possible batches that fit in GPU memory.

Utilizing the simulated workloads described in Sec \ref{subsec:model-description}, we estimate the effectiveness of the inference efficiency optimizations evaluated in Section \ref{subsec:model-description}. Based on these results, we select an inference framework with efficiency optimizations targeting large batch inference. Concretely, we consider inference with a dense model utilizing vLLM with CUDA graph serialization (eager mode off) on a single GPU and compare it to unoptimized inference native PyTorch as a lower bound on energy efficiency. In addition, we also model the idealized energy baseline based on the model and hardware configurations.

\paragraph{Classical NLP Tasks.}
\label{subsec:nlp-evaluation}

\begin{figure}
    \centering
    % \vspace{-2em}
    \includegraphics[width=0.85\linewidth]{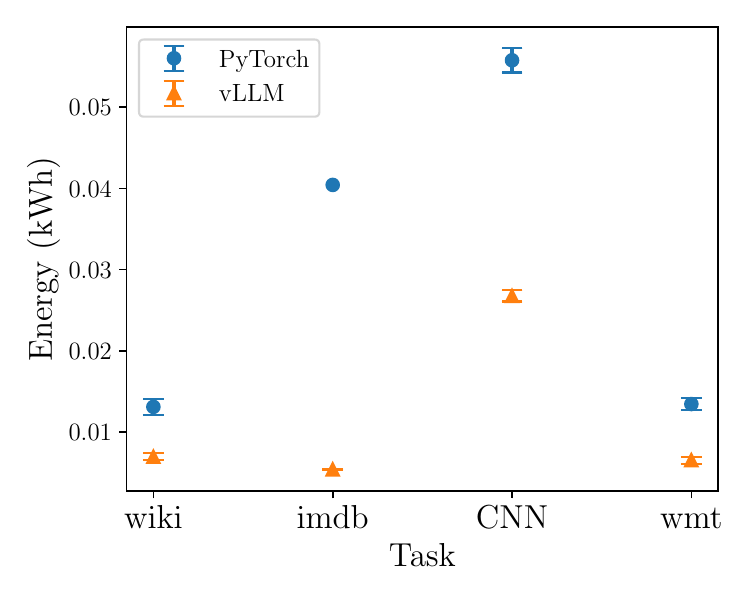}
    \vspace{-0.5em}
    \caption{\small Energy Comparison in doing inference over 1024 samples between PyTorch with Compilation off and vLLM with eager model off. }
    \vspace{-1
    em}
    \label{fig:energy-task-comparison}
\end{figure}

We benchmark the energy use in a set of classical natural language processing tasks in the English language: text classification (IMDB, \citealp{maas2011learning}), machine translation (WMT-14, \citealp{bojar-EtAl:2014:W14-33}), summarization (CNN-DailyMail, \citealp{nallapati2016abstractive}), and text generation (Wikitext-2 \cite{merity2016pointer}). 

For each of these tasks, we sample a subset of 1024 examples with statistics of each dataset 
% (mean, token length, maximum token lengths)
for the input and the output tokens provided in Table \ref{tab:nlp-token-stats}. We note that the input sequences were padded to the maximum sequence length. The energy profiles for the best run, characterized by the least energy are summarized in Figure \ref{fig:energy-task-comparison}, with consistent reductions in energy use provided by inference efficiency optimizations.

\paragraph{Real-World LLM Workloads}
\label{subsec:llm-evaluation}
Additionally, we estimate the energy intensity and effectiveness of efficiency optimizations on real-world LLM workloads. We simulate the offline processing of LLM inference requests as used in applications for short-form conversations with the Burst-GPT dataset \cite{wang2024burstgpt} and long context conversations and code completion with the Azure LLM Inference chat and code traces \cite{stojkovic2024dynamollm}. Each dataset provides a traces of LLM inference requests with their corresponding input context and output generation lengths.  As compared with the classical NLP tasks, modern LLM workloads tend to be longer in both input context and output generation token lengths, with code-assist applications having longer contexts, whereas conversational settings resulting in longer generations.

\begin{table}[ht]
\centering
\small
\begin{tabular}{lcc}
\toprule
Dataset & PyTorch \%$\Delta$ & vLLM \% $\Delta$ \\ \toprule
BurstGPT & 506.52\% & 63.75\% \\
Azure Code & 102.79\% & 26.59\% \\
Azure Conversation & 490.23\% & 64.22\% \\ \bottomrule
\end{tabular}
\caption{Percentage differences of energy consumption relative to theoretical values for Various Tasks with Offline Inference.}
\vspace{-5pt}
\label{tab:energy_percentage_diff}
\end{table}

Due to the larger number of requests and increased sequence lengths, we observe that these workloads require substantially larger amounts of energy. However, we find that proper applications of inference efficiency optimizations can substantially reduce energy costs with savings of 73.00\%, 37.58\%, and 72.18\% on BurstGPT, Azure Code and Conversation, respectively. 

\section{Related Work}

\label{sec:related-work}
% Efficient Inference
\paragraph{Efficient Methods for LLM Inference}
To meet the service-level-objective (SLO) serving requirements of real deployment settings, efficiency optimizations for LLM inference are often designed to optimize model serving speed, as measured by latency and time-to-first-token. A variety of methods have been developed to meet these latency constraints, including: continuous batching \citep{yu2022orca}, model parallelism \citep{narayanan2021efficient,huang2019gpipe,li2020pytorch}, speculative decoding \citep{liu2024optimizing,leviathan2023fast,chen2023accelerating,chen2024magicdec}, and disaggregated serving \citep{zhong2024distserve}.

Solely optimizing system performance for speed is insufficient in characterizing and does not provide insight into the model energy use and resulting carbon emissions of LLM inference; as such methods may require additional computation or exhibit low correlation between efficiency cost indicators \citep{dehghaniefficiency}. Recent work has explored methods for explicitly reducing energy requirements and carbon emissions for LLM serving via disaggregated serving over heterogeneous hardware \citep{shi2024greenllm}, system-wide scheduling and request routing to energy-optimized instances \citep{stojkovic2024dynamollm}, and prompt directives to induce shorter sequence generations \citep{li2024sprout}. However, the exact impact or improvements in energy requirements for latency-optimized methods remains not fully characterized.

\paragraph{Estimations and Measurement of of Energy Use in NLP}
The energy and carbon emissions of machine learning models have been a growing concern in the research community and industry as the scale of models and prevalence of deployment has increased \citep{schwartz2020green,wu2022sustainable}. Estimations of the energy requirements and environmental impact of LLMs has largely focused on estimation of costs for pretraining and finetuning due to the large singular costs of model developments \citep{strubell2020energy, wang2023energy, luccioni2023estimating,faiz2023llmcarbon}; with large industrial developers similarly reporting the energy required for pretraining \citep{olmo20242, morrison2025holistically, dubey2024llama}.

In contrast to training, inference workloads are higher in variability with variation in request frequencies, batching, input and output sequence lengths executed over diverse hardware platforms at scale; and more complex energy use profiles due to variations in power draw during prefill and decoding stages of generation \citep{patel2024characterizing}. Previous work has investigated the comparative energy cost of machine learning models across various tasks \citep{luccioni2024power,luccioni2024light}, the energy costs of LMs of various sizes \citep{samsi2023words,wu2025unveiling}, the effects of hardware configurations (i.e. GPU power capping and frequency scaling; \cite{samsi2023words}), and the impact of sequence length variability and batching strategies \cite{patel2024characterizing, stojkovic2024towards,wilkins2024offline}. However, such evaluations of inference energy use often rely on simplified deployment settings with limited sets of model architectures and serving frameworks.

\section{Conclusion}
\label{sec:conclusion}

In this work, we evaluate the impact of common inference efficiency optimizations on the energy requirements of large language model serving. We examine a variety of optimization techniques and evaluate on representative data corresponding to classical NLP tasks as well as modern LLM deployment settings. We conclude that the effectiveness of latency optimizations in reducing energy use is highly sensitive to the shape of the input data, underlying hardware architecture, and software framework implementations; and that optimizations cannot be applied uniformly. 

Additionally, we conduct a case study of classical NLP tasks and real-world LLM inference workloads and find that proper application of the studied inference optimizations can reduce total energy use by up to 73\% on the BurstGPT chat dataset.

\section*{Limitations and Risks}
In this work, we evaluate the energy efficiency and carbon emissions of LLM inference as approximated by total GPU power usage. Although GPUs the majority of arithmetic operations required for inference and operate at a higher TDP than other components, we do not account for the energy use by other other components of the hardware system such as power use from CPU, memory, or disk storage \cite{mcallister2024call,patel2024characterizing}; or estimate the energy requirements of other hardware accelerator architectures (e.g. TPUs, NPUs, etc.).  Likewise, we conduct an investigation of commonly used inference software frameworks and standard efficiency optimizations. However, there remain other settings and computational optimizations that can be applied to LLM inference, such as utilizing: reduced or mixed precision,  adaptive adjustment of GPU frequency, additional forms of model parallelism, or other forms of load management and workload scheduling; which remain out of the scope of this work \cite{stojkovic2024dynamollm}.

In this work, we primarily focus on the operation energy use of machine learning inference. Estimation of the embodied costs of inference; and the costs of machine learning training remain out of the scope of this work.

Although improved characterization of the energy use of LLM inference can be used to design more efficient serving settings and reduce the energy needs of inference, it is possible that reductions in the cost of pretraining may then lead more individuals and organizations to pursue large model pretraining (i.e. Jevons Paradox).

\bibliography{latex/custom}
\appendix
\appendix
\section{Hardware Details}
\label{appx:hardware}
In Table \ref{tab:hardware}, we provide additional details on the hardware configurations of the nodes used in our benchmarking experiments.

\begin{table*}[h]
    \centering
    \footnotesize
    \begin{tabular}{cccccc}
    \toprule
        \textbf{CPU} & \textbf{RAM} & \textbf{GPU} & \textbf{GPU TDP} & \textbf{FP32 TFLOPS} & \textbf{Bfloat16 TFLOPS}\\
    \midrule
        256xAMD EPYC 7763 &  1TB    & Nvidia RTX A6000      & 300W & 38.7 & -- \\
        128xAMD EPYC 7513 & 500GB   & Nvidia RTX A6000 Ada  & 300W & 91.1 & -- \\
        128xAMD EPYC 7763 &  1TB    & Nvidia RTX A100-80 GB & 300W & 156 & 312 \\
    \bottomrule
    \end{tabular}
    \caption{Node Hardware Specifications}
    \label{tab:hardware}
\end{table*}

\section{Dataset Licenses}
\label{appx:licenses}
The CNN-DailyMail dataset used for summarization is released under the Apache-2.0 License. The dataset Wikitext-2 dataset for text generation is available under the Creative Commons Attribution-ShareAlike License. The WMT-14 translation datasets are released for non-commercial use. 
 The BurstGPT and Azure trace datasets are released under CC-BY-4.0 licenses.
 
\section{Acknowledgment of AI Assistance }
\label{appx:ai-assist}
Artificial intelligence assistance was used to assist in literature review and for code completion assistance, specifically during the creation of visualizations.

\section{Additional Optimzations: Continuous Batching}
\label{appx:software}
\begin{figure*}
    \centering
    \begin{subfigure}{0.3\textwidth}
        \centering
        \includegraphics[width=\linewidth]{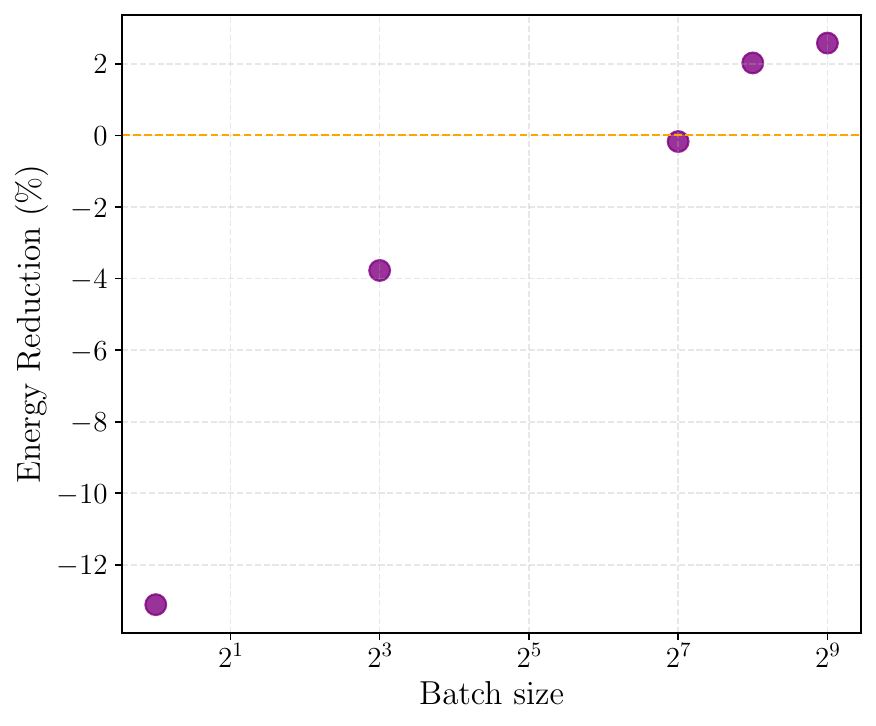}
        \caption{A100 80GB PCIe}
    \end{subfigure}
    \hfill
    \begin{subfigure}{0.3\textwidth}
        \centering
        \includegraphics[width=\linewidth]{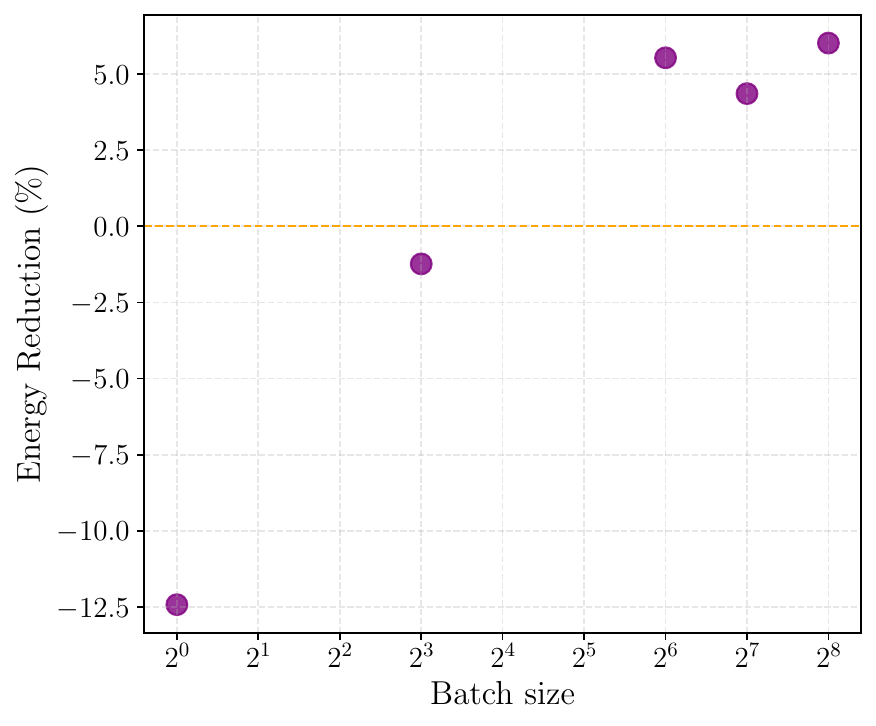}
        \caption{A6000 Ada}
    \end{subfigure}
    \hfill
    \begin{subfigure}{0.3\textwidth}
        \centering
        \includegraphics[width=\linewidth]{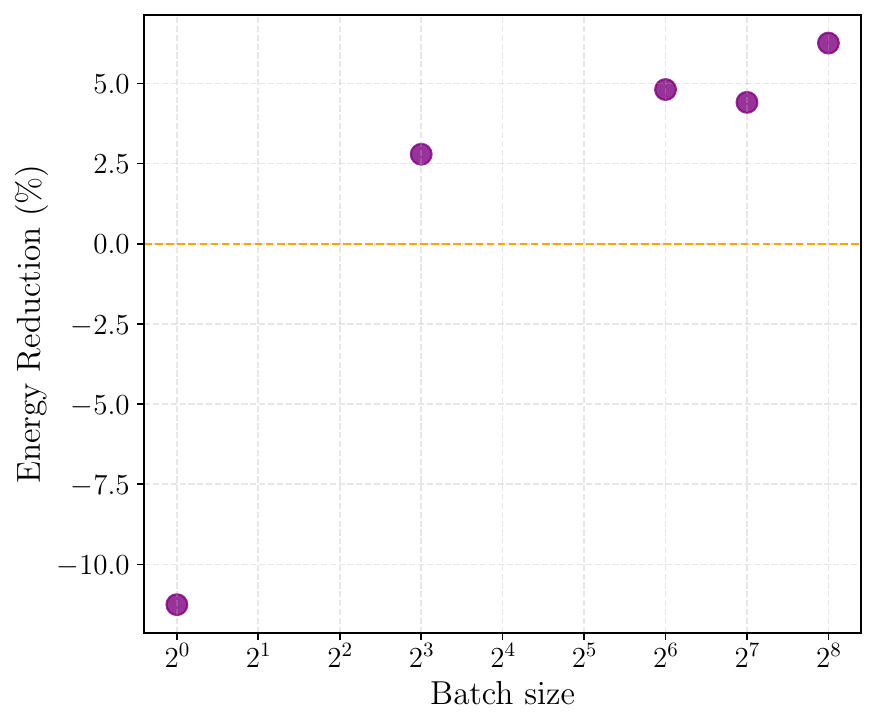}
        \caption{A6000}
    \end{subfigure}
    
    \caption{\textbf{Energy reduction comparison between online and offline serving modes across different GPUs $(E_{offline} - E_{online})*100/E_{offline})$.} The optimizations employed for online serving save up to 5\% energy at larger batch sizes}
    \label{fig:energy-reduction}
\end{figure*}

In Figure \ref{fig:energy-reduction}, we present additional results on the impact of vLLM's continuous batching for online inference in which we observe that at large batch sizes continuous batching yields reductions in energy use. 

\section{Additional Sequence Length Results}
\label{appx:seqlen}
\begin{figure*}[ht!]
        \centering
        \includegraphics[width=0.31\textwidth]{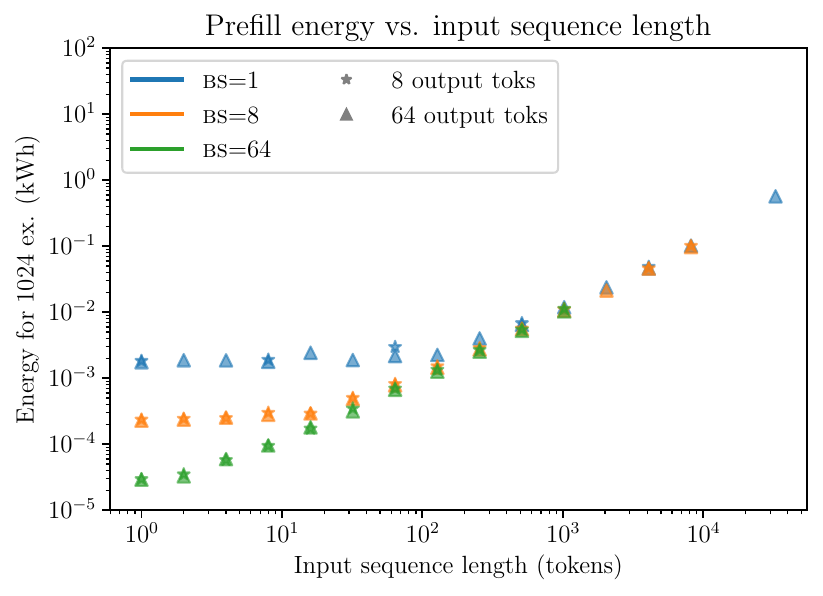}
        \includegraphics[width=0.31\textwidth]{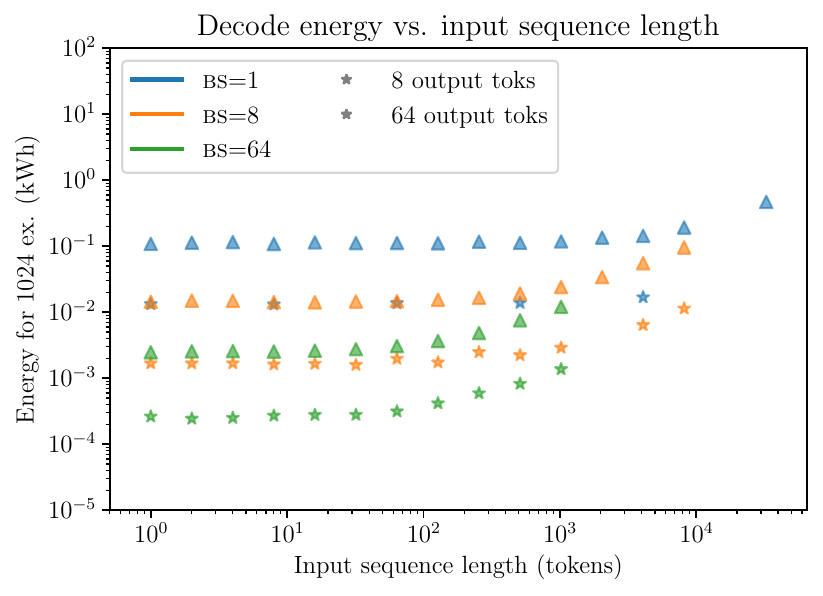}
        \includegraphics[width=0.31\textwidth]{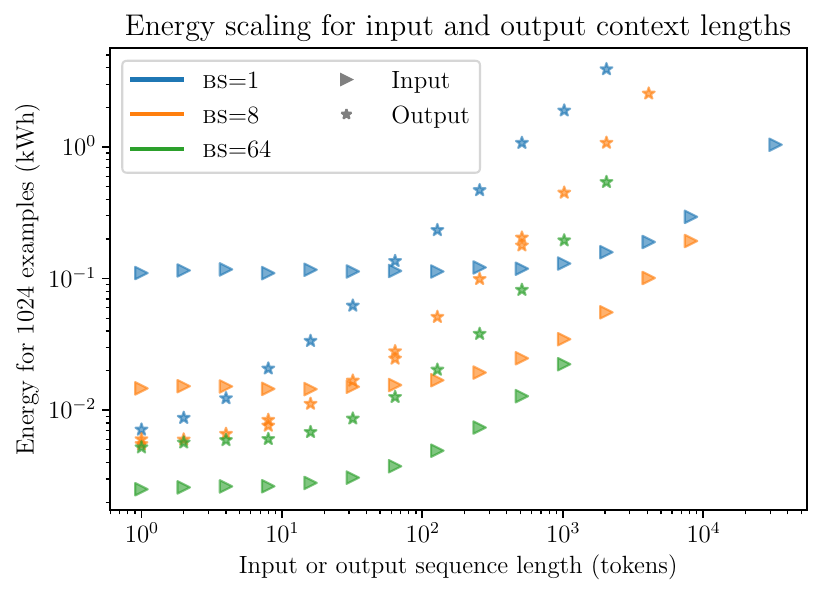}
        \caption{
            Controlled sweeps of input and output sequence lengths on A6000 GPUs, with vanilla PyTorch backend. 
        }
        \label{fig:inout-seqlen-pytorch}
    \end{figure*}

In Figure \ref{fig:inout-seqlen-pytorch}, we present additional results on the effects of scaling input and output sequence lengths with the PyTorch framework. 

\begin{figure*}
\centering
    \includegraphics[width=0.4\textwidth]{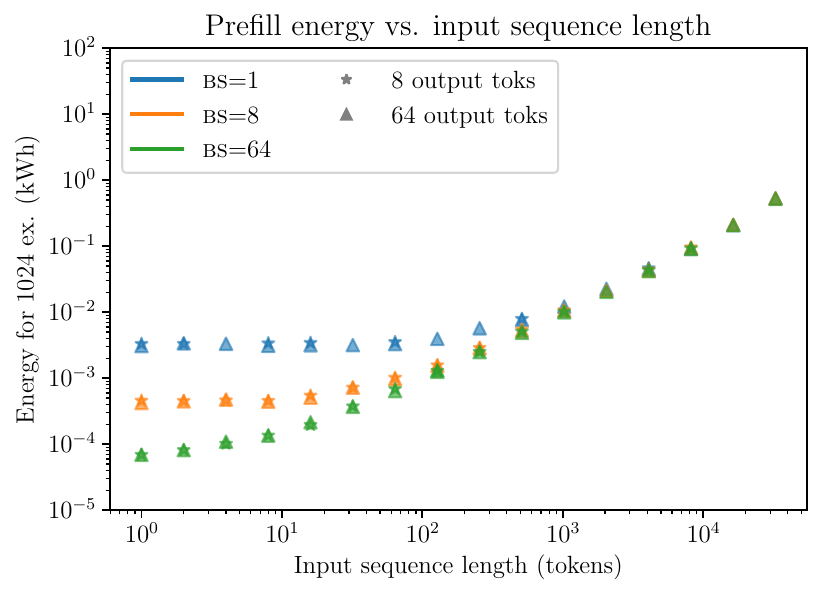}
    \includegraphics[width=0.4\textwidth]{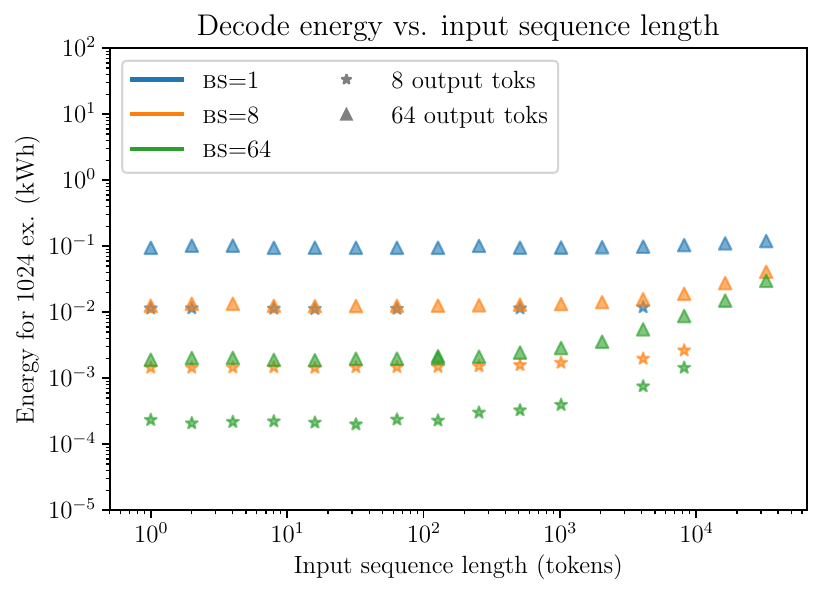}
    \includegraphics[width=0.4\textwidth]{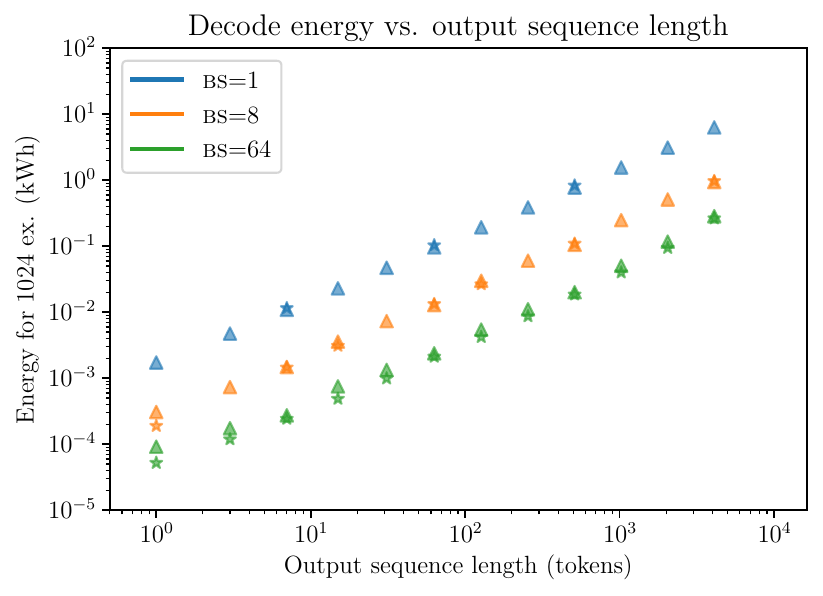}
    \includegraphics[width=0.4\textwidth]{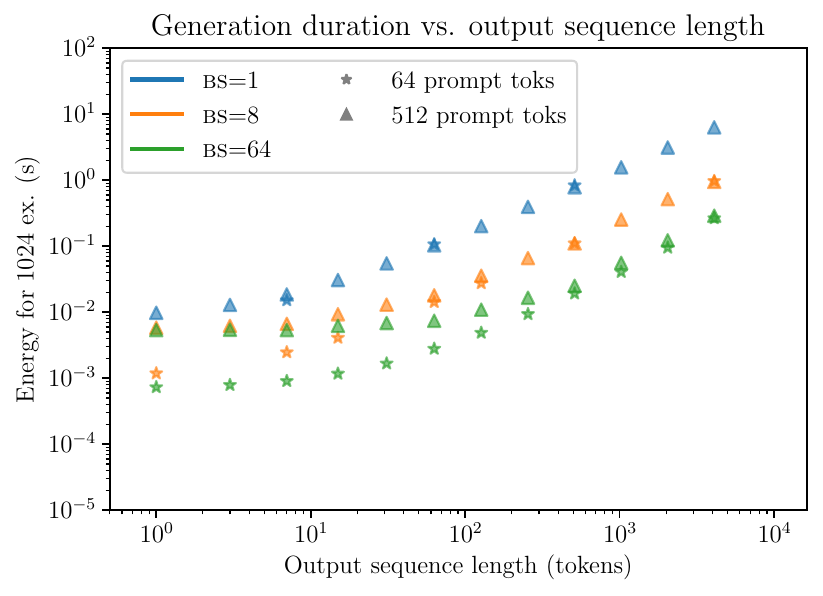}
    \caption{Controlled sweeps of input and output sequence lengths on A6000 GPUs, with vLLM offline inference. Here, we display multiple fixed sequence length sizes for comparison as we sweep across batch size and the other dimension of sequence length.
    }
    \label{fig:inout-seqlen-compare}
\end{figure*}

\begin{figure}
    \centering
    \includegraphics[width=0.8\linewidth]{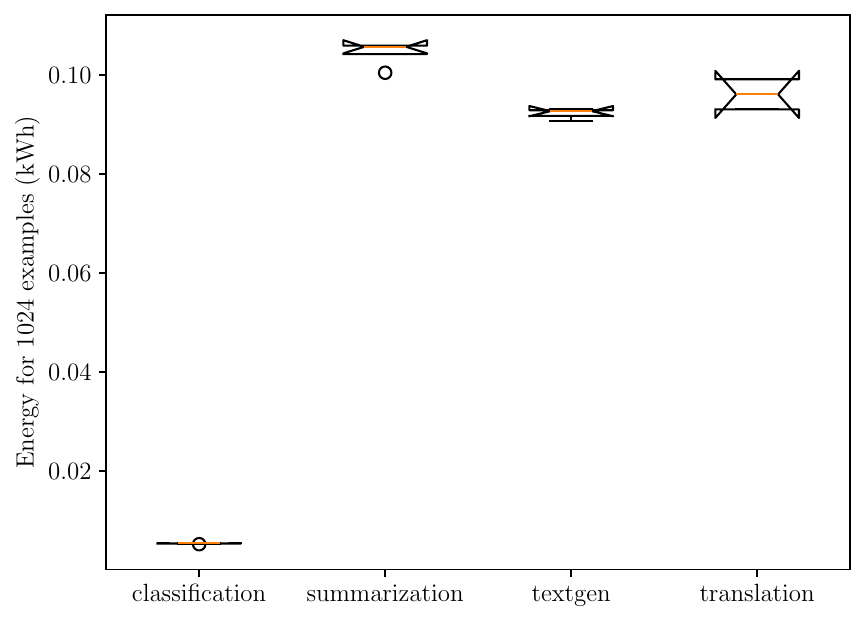}
    \includegraphics[width=0.8\linewidth]{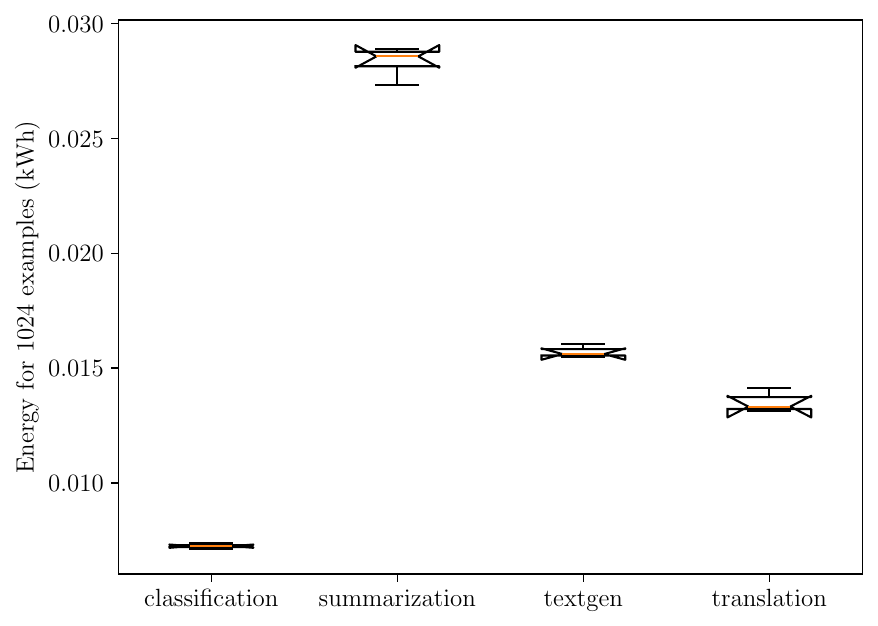}
    \includegraphics[width=0.8\linewidth]{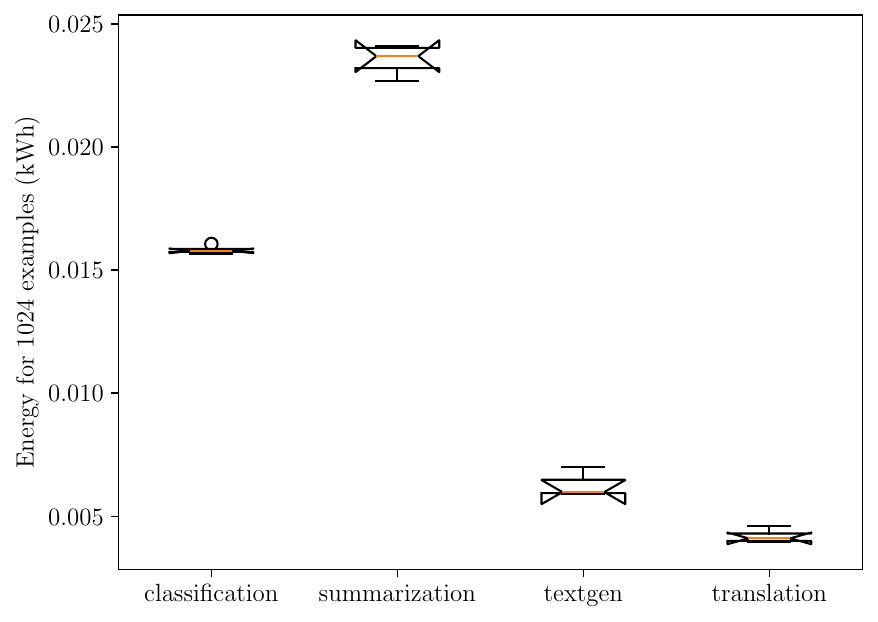}
    \caption{Classical NLP tasks and their energy intensities with vLLM backends. From top to bottom, the batch size varies from 1, 8, to 128}
    \label{fig:tasks-vllm}
\end{figure}

\end{document}